\documentclass{llncs}
\usepackage[dvips]{graphicx}
\usepackage{algorithm} 
\usepackage{algorithmic}  
\begin{document}

\title{Mining Patterns with a Balanced Interval}
\author{Edgar H. de Graaf, Joost N. Kok \and Walter A. Kosters}
\institute{\textit{Leiden Institute of Advanced Computer Science\\
Leiden University, The Netherlands} \\
\texttt{edegraaf@liacs.nl} }

\pagestyle{empty}
\maketitle
\thispagestyle{empty}
\begin{abstract}
In many applications it will be useful to know
those patterns that occur with a balanced interval, e.g., a certain 
combination of phone numbers are called almost every Friday or a
group of products are sold a lot on Tuesday and Thursday.

In previous work we proposed a new measure of support 
(the number of occurrences of a pattern in a dataset), where we count 
the number of times a pattern occurs (nearly) in the middle between 
two other occurrences.  If the number of non-occurrences between 
two occurrences of a pattern stays almost the same then we call
the pattern balanced.

It was noticed that some very frequent patterns obviously also
occur with a balanced interval, meaning in every transaction. 
However more interesting patterns might occur, e.g., every three transactions.
Here we discuss a solution using standard deviation and average. Furthermore
we propose a simpler approach for pruning patterns with a balanced interval,
making estimating the pruning threshold more intuitive.
\end{abstract}

\section{Introduction}

% * Vertel iets generieks over onze oplossing *
Mining frequent patterns is an important area of data mining
where we discover substructures that occur often in (semi-)structured data.
In this work we will further investigate one of the simplest structures:
itemsets. However the principles of  balanced patterns are easily extended to sequential pattern
mining, tree and graph mining.
In earlier work we proposed an algorithm that discovers \emph{stable patterns} that occur
at regular moments, or rather in regular intervals, enabling us to mine for 
events that occur, e.g., every Friday. In this work we will introduce a new
approach to mining for patterns with a stable interval. Note that the transactions
in this paper have an order.
In order to distinguish it from stable patterns we will call these new patterns
\emph{balanced patterns}. 
With this new approach we will offer solutions for problems in 
our work done in \cite{GraafEH2}:

\begin{itemize}
\item Patterns occurring in every transaction made it hard to discover patterns
with a more interesting intermediate interval.
\item The threshold for pruning was a certain value that a measure for stability 
needed to achieve. Even though a formula was given to estimate this value, an
easily understandable value was lacking.
\end{itemize}

In Section 2.1 we will repeat some important definitions to make this work self contained, 
however in depth information can be found in \cite{GraafEH2}.

We will define our approach to mining balanced patterns and show its usefulness. 
To this end, this paper makes the following contributions:
\\
--- We will \textbf{define balanced patterns and show their use}. 
These balanced patterns will enable the user to better 
filter uninteresting patterns (Section 2).
\\
--- Furthermore we will \textbf{propose an algorithm} that will enable us to mine
balanced patterns (Section 3).
\\
--- Finally we will empirically show that 
\textbf{the algorithm can find interesting patterns} efficiently (Section 4).

% * Leg het mine van websites probleem uit *

A typical example is the mining of an access log from the Computer Science 
department of Leiden University. This access log will first
be converted to sets of properties we are interested in, e.g.,
pages visited every half-hour. From here on we call this dataset the 
\emph{website} dataset.

%* Ander dataset is trading agent server? *

%* Het werk van andere, related work *

This research is related to work done on the (re)definition
of support, using time with patterns and the incorporation of 
distance measured by the number of transactions between pattern
occurrences.
The notion of support was first introduced by Agrawal et al. in
\cite{apriori} in 1993. Since then many new and faster algorithms
where proposed. We make use of \textsc{Eclat}, developed by Zaki et al.
in \cite{Zaki1}. Steinbach et al. in
\cite{SteinbachM} generalized the notion of support providing a
framework for different definitions of support in the future. Our
work is also related to work described in \cite{LiY} where association
rules are mined that only occur within a certain time interval. Furthermore
there is some minor relation with mining data streams as described in \cite{ChenY,GiannellaC,TengW},
in the sense that they use time to say something about the importance of a pattern.

Finally this work is related to some of our earlier work.
Results from \cite{GraafJ1} indicated that 
the biological problem could profit from
incorporating consecutiveness into frequent itemset mining,
which was elaborated in~\cite{GraafGraaf}.
In the case of stable patterns we also make use of the
transactions and the distance between them.
Secondly in \cite{GraafEH} it was mentioned that
support is just another measure of saying how good a pattern fits
with the data. There we defined different variations of this
measure, and stability can been seen as one such variation.
Stable patterns and an algorithm to discover them are
defined in \cite{GraafEH2}.

\section{Regular Occurrence}

%* Defineer Stable Patterns *
In this section we will repeat the definition of stable patterns to better
understand the problems and the difference with the definition of balanced
patterns. In particular,
patterns that occur at regular intervals
(e.g., at equidistant time stamps) will be called stable or balanced.
In the case of \emph{stable patterns}, in order to judge this property, we will determine how often events 
occur ``in the middle'' between two other events \cite{GraafEH2}. In the case of \emph{balanced patterns}
we prune patterns that do not have at least one frequent intermediate distance (between all occurrences) 
and we filter those patterns that have a too high deviation for all distances 
between successive occurrences. Furthermore we filter patterns that do not 
reach a certain minimal average distance for all successive occurrences.

\subsection{Stable Patterns}

In this paper a dataset consists of 
transactions that take zero time.
Each transaction is an itemset, i.e., a subset of
$\{1,2,3,\ldots,\mathit{max}\}$ for some fixed integer
$\mathit{max}$. The transactions can have time stamps;
if so, we assume that the transactions take place at different moments.
We choose some notion of \emph{distance} between transactions;
examples include: (1) the distance is the time between the two
transactions and (2) the distance is the number of transactions
(in the \emph{original} dataset) strictly in between the two transactions.
In this paper we will use (2) in all our examples.
We will define $\mathit{Trans}(p)$ as the series of transactions
that contain pattern (i.e., itemset) $p$;
the \emph{support} of a pattern $p$ is the number
of elements in this ordered series.

We now define \emph{$w$-stable patterns} as itemsets that occur frequent
(support $\geq \mathit{minsup}$) in the dataset and that 
have \emph{stability value} $\geq \mathit{minstable}$,
where the values $\mathit{minsup}$ and
$\mathit{minstable}$ are user defined thresholds.
A \emph{$w$-good triple} $(L,M,R)$ consists of three transactions
$L$, $M$ and $R$, occurring in this order, such that
$|\mathit{distance}(L,M)-\mathit{distance}(M,R)|\leq2\cdot w$;
here $w$ is a pregiven small constant $\geq 0$, e.g., $w=0$.
The stability value of a pattern $p$ is the number of 
$w$-good triples in $\mathit{Trans}(p)$, plus the number of transactions
in $\mathit{Trans}(p)$ that occur as left endpoint in a $w$-good triple,
plus the number of transactions
in $\mathit{Trans}(p)$ that occur as right endpoint in a $w$-good triple.

Note that the stability value of a pattern $p'$ with $p'\subseteq p$
is at least equal to that of $p$: the so-called \textsc{Apriori}
or anti-monotone property. Also note that the stability value
remains the same if we consider the dataset in reverse order.

In our work on stable patterns \cite{GraafEH2} we showed that equidistant events 
are ``very'' stable (in case $w=0$).

\begin{example}
Suppose we have the following itemsets in our dataset:

\medskip
\begin{tabular}{l@{\ \ \ \ \ }l}
\ \ \ \ transaction 1: \{$A$, $B$, $C$\}\\
\ \ \ \ transaction 2: \{$D$, $C$\}\\
\ \ \ \ transaction 3: \{$A$, $B$, $E$\}\\
\ \ \ \ transaction 4: \{$E$, $F$\}\\
\ \ \ \ transaction 5: \{$A$, $B$, $F$\}\\
\ \ \ \ transaction 6: \{$E$, $F$\}\\
\ \ \ \ transaction 7: \{$A$, $B$, $F$\}\\
\ \ \ \ transaction 8: \{$E$, $F$\}\\
\ \ \ \ transaction 9: \{$A$, $B$, $C$\}\\
\end{tabular}

\end{example}
The stability value (with $w=0$) of \{$A$,$B$\} is 
$4+3+3=10$, the maximal value possible.
There are 4 0-good triples; we have 3 transactions that are left (right) endpoint 
of a 0-good triple (see picture below, left).
If we insert two transactions \{$E$, $F$\} between transaction
1 and 2, and also two between 8 and 9, we still have 
4 0-good triples, but now we only have 
2 transactions that are left (right) endpoint
of a good 0-triple (see picture below, right), leading to stability value $4+2+2=8<10$.
This example shows that in order to guarantee equidistance one has to add left and
right endpoints to the stability value. 
%Basically if many different occurrences are 'used' as
%left and right endpoint than more different points are stable relative to each other and the corresponding
%stability value should be higher.
\begin{figure}[!ht]
\begin{center}
\begin{picture}(245,15)
\put(0,8){\line(1,0){90}}
\put(5,8){\circle*{4}}
\put(15,6){\line(0,1){4}}
\put(25,8){\circle*{4}}
\put(35,6){\line(0,1){4}}
\put(45,8){\circle*{4}}
\put(55,6){\line(0,1){4}}
\put(65,8){\circle*{4}}
\put(75,6){\line(0,1){4}}
\put(85,8){\circle*{4}}

\put(120,8){\line(1,0){130}}
\put(125,8){\circle*{4}}
\put(135,6){\line(0,1){4}}
\put(145,6){\line(0,1){4}}
\put(155,6){\line(0,1){4}}
\put(165,8){\circle*{4}}
\put(175,6){\line(0,1){4}}
\put(185,8){\circle*{4}}
\put(195,6){\line(0,1){4}}
\put(205,8){\circle*{4}}
\put(215,6){\line(0,1){4}}
\put(225,6){\line(0,1){4}}
\put(235,6){\line(0,1){4}}
\put(245,8){\circle*{4}}
\end{picture}
\end{center}
\vspace{-10mm}
\end{figure}

\subsection{Balanced Patterns}

In this section we will define balanced patterns. We first discuss
several problems and possibilities, and finally give the proper definition.
We call the occurrences balanced if between two successive occurrences
there is (almost) always the same amount of transactions. 
%This basically is equal to the definition of a stable interval.

The problem with patterns with balanced occurrences is that 
an itemset may occur less balanced than a superset of this itemset. 
Patterns occurring with a balanced interval do not have 
the \emph{anti-monotone property}, where the subset is either equally
good or better than the superset. In the balanced pattern case: the subset is not
always more (or equally) balanced than the superset.
This value will be used for pruning.

\begin{example}
Say that item $A$ occurs in transactions 1, 4, 7 and 10 and item $B$
occurs in transaction 4, 7, 10 and 13 then the itemset $\{A, B\}$ will
occur in transaction 4, 7 and 10. Both $A$ and $B$ have three times
two transactions between occurrences (successive and non-successive). However $\{A, B\}$ has only
two times two transactions between occurrences because an occurrence can
only become a non-occurrence and not the other way around.
\end{example}

For our definition of balanced patterns we first notice that all
balanced occurrences (successive and non-successive) should 
have at least one intermediate distance a minimal number of times.
Furthermore if you count the distances \emph{between all occurrences} then this count is 
anti-monotone: a superset never has more of one particular distance. 
This is obvious because the number of occurrences will never increase for a superset 
and as a consequence the count of one particular distance will never increase.
This property is also anti-monotone if we limit the distances we count, 
e.g., we count a distance only if it is smaller than 10 in-between transactions.

\begin{example}
The following table, where we only count upto 4 in-between transactions,
is an example of counting the distances:\\

\begin{center}
\begin{tabular}{|c|c|}
\hline
In-between Transactions & Count \\
(Distance) & \\
\hline
0 & 0 \\
\hline
1 & 5 \\
\hline
2 & 200 \\
\hline
3 & 30 \\
\hline
4 & 199 \\
\hline
\end{tabular}
\end{center}
\noindent The \emph{balanced value} for the pattern with these counts will be 200, the highest count in the table.
\end{example}

Still if we only look at the distance count we will not find the balanced patterns
we want, since patterns that occur with very unbalanced intervals might still have a minimum
amount of one particular distance. We filter those patterns by keeping the distance
between occurrences that immediately succeed each other (instead of taking all distances). 
If a pattern is balanced then these distances should approach the average of all these
distances. Their standard deviation will be near $0$, since one distance should occur the most.
Note that in calculating the standard deviation we do not limit the distances we consider.
This can be done because the number of possible distances is far less for
successive occurrences.

Now we can find all balanced patterns, however we will still find many patterns
that are occurring every transaction. Their distance is almost always 0 and although
they are well balanced they are often not interesting. These patterns can be filtered
if we demand a certain average distance, e.g., if the user-defined threshold 
$\mathit{minavg}$ is set to $1$ then all these patterns will be filtered out, since their
average distance approaches $0$. 

\bigskip
\noindent
The definition of balanced patterns should be the following:
A pattern is called a \emph{balanced pattern} if among all 
occurrence pairs there is a distance that occurs at least a user-defined number of times
($\mathit{minnumber}$) and the distance between successive occurrences have maximally a 
user-defined standard deviation ($\mathit{maxstdev}$) and minimally a user-defined average ($\mathit{minavg}$).
%\bigskip

\section{Algorithm}

We now consider algorithms that find all frequent itemsets, given a
database. A \emph{frequent} itemset is an itemset with support at
least equal to some pre-given threshold, the so-called
\emph{minsup}. Thanks to the \textsc{Apriori} property many efficient algorithms exist. However,
the really fast ones rely upon the concept of \textsc{FP-tree} or
something similar, which does not keep track of in-between distances. This makes
these algorithms hard to adapt for use in balanced patterns. 

One fast algorithm that does not make use of \textsc{FP-tree}s is called
\textsc{Eclat} \cite{Zaki1}. \textsc{Eclat} grows patterns recursively while
remembering which transactions contained the pattern, making it very suitable 
for balanced patterns. In the next recursive
step only these transactions are considered when counting the occurrence of
a pattern. All counting is done by using a matrix and patterns are extended with
new items using the order in the matrix. This can easily be adapted to incorporate 
balance counting.

Our algorithm \textsc{BalanceClat} will use the \textsc{Eclat} algorithm. However instead of counting
support we count the different distances between all occurrences, e.g., pattern $A$
has 10 times 3 transactions between occurrences.
We will prune on this value instead of pruning on the minimal support threshold.
In this case the user-defined threshold will be the minimal number of times
at least one of $\ell + 1$ distances $\{0, 1, 2, \ldots, \ell\}$ is seen. 
For balanced patterns we consider this threshold to be the
$\mathit{minnumber}$ threshold. As said before, we can only find balanced patterns
if we also demand a maximal standard deviation for distances between occurrences. 
This will be done by introducing the $\mathit{maxstdev}$ threshold. Finally we are
not interested in patterns occurring in every transaction. We introduce a third
user-defined threshold that demands a minimal average distance: $\mathit{minavg}$.
For $\mathit{maxstdev}$ and $\mathit{minavg}$ we only use distances between successive occurrences
and for $\mathit{minnumber}$ all distances $\leq \ell$.

%Suppose items are from the set
%$\mbox{$\cal{I}$}=\{1,2,\ldots,n\}$, where $n\geq 1$ is a fixed
%integer constant. A \emph{transaction} is an \emph{itemset}, which is a
%subset of $\cal{I}$. A \emph{database} is an \emph{ordered series}
%of $m$ transactions, where $m\geq 1$ is a fixed integer constant.
%If an itemset is an element of a database, it is usually referred
%to as a transaction.

%The \emph{traditional support} of an itemset $I$ with respect to a
%database $\cal{D}$, denoted by
%$\mathrm{TradSupp}(I,\mbox{$\cal{D}$})$, is the number of
%transactions from $\cal{D}$ that contain $I$. Clearly,
%$0\leq\mathrm{TradSupp}(I,\mbox{$\cal{D}$})\leq m$.

We now propose a more general definition. Suppose we have an itemset
$I$ and let $O_j\in\{0,1\}$ ($j=1,2,\ldots,r$) denote whether or not
the $j^{\mathrm{th}}$ transaction in some subset $\cal{S}$  
of the database $\cal{D}$ contains $I$ ($O_j$ is 1 if it does contain $I$, and 0 otherwise;
the $O$'s are referred to as the \emph{$O$-series}), $r = |\cal{S}|$. 
The function $\varphi: N \rightarrow N$
is a translation from the index $j$ for the $j$-th transaction in $\cal{S}$ to the index
$k$ giving the position of the same transaction in $\cal{D}$.

The main adaptation to \textsc{Eclat} is replacing support with a \emph{balance value}
denoted with $t$. Also it calculates the standard deviation ($\mathit{stdev}$) 
and average distance ($\mathit{avgdist}$) for the successive occurrences:

\begin{tabbing}
XXX\=XX\=XX\=XX\=XX\=XX\=XX\=\kill
\>$j := 2$, $h:=-1$\\
\>$\mathit{succdists} :=$ sequence of distance counts between \emph{successive} occurrences\\
\>$\mathit{alldists} :=$ sequence of distance ($\leq \ell$) counts between \emph{all} occurrences\\
\>{\bf while} $(\ j \leq r\ )$ {\bf do}\\
\>\>{\bf if} $(\ O_j=1\ )$ {\bf then}\\
\>\>\>$i := 1$\\
\>\>\>{\bf while} $(\ i < j\ )$ {\bf do}\\
\>\>\>\>{\bf if} $(\ O_i=1 \mathbf{\ and\ } \varphi(j) - \varphi(i)-1 \leq \ell \ )$ {\bf then}\\
\>\>\>\>\>$\mathit{alldists}_{\varphi(j) - \varphi(i)-1} := \mathit{alldists}_{\varphi(j) - \varphi(i)-1} + 1$\\
\>\>\>\>{\bf fi}\\
\>\>\>\>$i := i + 1$\\
\>\>\>{\bf od}\\
\>\>\>{\bf if} ($\ h \neq -1\ $) {\bf then}\\
\>\>\>\>$\mathit{succdists}_{\varphi(j) - \varphi(h)-1} := \mathit{succdists}_{\varphi(j) - \varphi(h)-1} + 1$\\
\>\>\>{\bf fi}\\
\>\>\>$h := j$\\
\>\>{\bf fi}\\
\>\>$j := j + 1$\\
\>{\bf od}\\
\>$t:= \mathit{max}(\mathit{alldists})$, the largest count in the sequence\\
\>$\mathit{stdev} :=$ standard deviation for $\mathit{succdists}$\\
\>$\mathit{avgdist} :=$ average for $\mathit{succdists}$, also denoted with $\mathit{avg}$($\mathit{succdists}$)\\
\end{tabbing}

\noindent The standard deviation for $\mathit{succdists}$ can simply 
be calculated in the following way:

\begin{equation}\label{een}  
	\sqrt{ \textstyle{\sum _{i}} (\mathit{avg}(\mathit{succdists}) - i)^2 \cdot \mathit{succdists}_{i}\; / \;\sum_{i} \mathit{succdists}_{i} }
\end{equation}

\textsc{Eclat} can now prune using the balance value $t$ (if $t < \mathit{minnumber}$) and patterns
are only displayed if their standard deviation and average distance are
sufficient. These are straightforward adaptations that will not be given in detail.
%In some cases it might be preferable to skip certain distances all together, e.g.,
%we don't count the 0-distances. These can be skipped by not counting $\mathit{dists}_{0}$,
%in this way we can filter out patterns that have only many times the 0-distance.
%Note that $\mathit{succdists}_{0}$ needs to be counted else it can not be guaranteed
%that a pattern with a low standard deviation for its interval also is balanced.

Standard deviation changes if patterns occur less balanced in a certain small number 
of successive transactions, small periods. In some cases it might be preferable to
remove the influence of these periods. One possible approach is to calculate
average distance and the standard deviation for \emph{frequent distances} (for successive occurrence)
only. The value for filtering with standard deviation 
for the sequence $\cal{Q}$ $= \langle y|y = \mathit{succdist_i}, y \geq \mathit{mindistfreq}\rangle$ will be:

\begin{equation}\label{twee} 
	\mathit{stdev} = \left\{ \begin{array}{l l} \sqrt{\sum_{i} (\mathit{avg}(\mathit{\cal{Q}}) - i)^2 \cdot \mathit{\cal{Q}}_{i}\;/\;\sum_{i} \mathit{\cal{Q}}_{i} } & \quad \mbox{if $\mathit{\cal{Q}}$ is \emph{not} empty}\\
	\mathit{maxstdev} + 1 & \quad \mbox{otherwise}\\
	 \end{array} \right.
\end{equation}

\noindent Note that via the threshold $\mathit{mindistfreq}$ the user decides when a distance is considered frequent.

\section{Results and Performance}

% * Omschrijf puntgewijs waarom deze experimenten

The experiments were done for three main reasons. First of all
we want to show \emph{known balanced patterns will be found} also
in the case of noise. 
Secondly we want to show that \emph{interesting balanced patterns can be found}
in real datasets. Finally we want to  \emph{show runtime for real data} and
\emph{how the $\mathit{minnumber}$ threshold influences runtime}. 
%The threshold used for pruning
%in the case of stable patterns is different, however we will make a comparison
%between runtime.
% * Waarom de datasets?

Our implementation of the balanced pattern mining algorithm is called \textsc{BalanceClat}.
All experiments were performed on an Intel Pentium 4 64-bits 3.2 GHz machine with 3 GB memory. As operating system Debian
Linux 64-bits was used with kernel 2.6.8-12-em64t-p4.

\begin{figure}[!ht]
\hfill
\begin{minipage}[t]{.49\textwidth}
\begin{center}
\includegraphics[width=6.0cm]{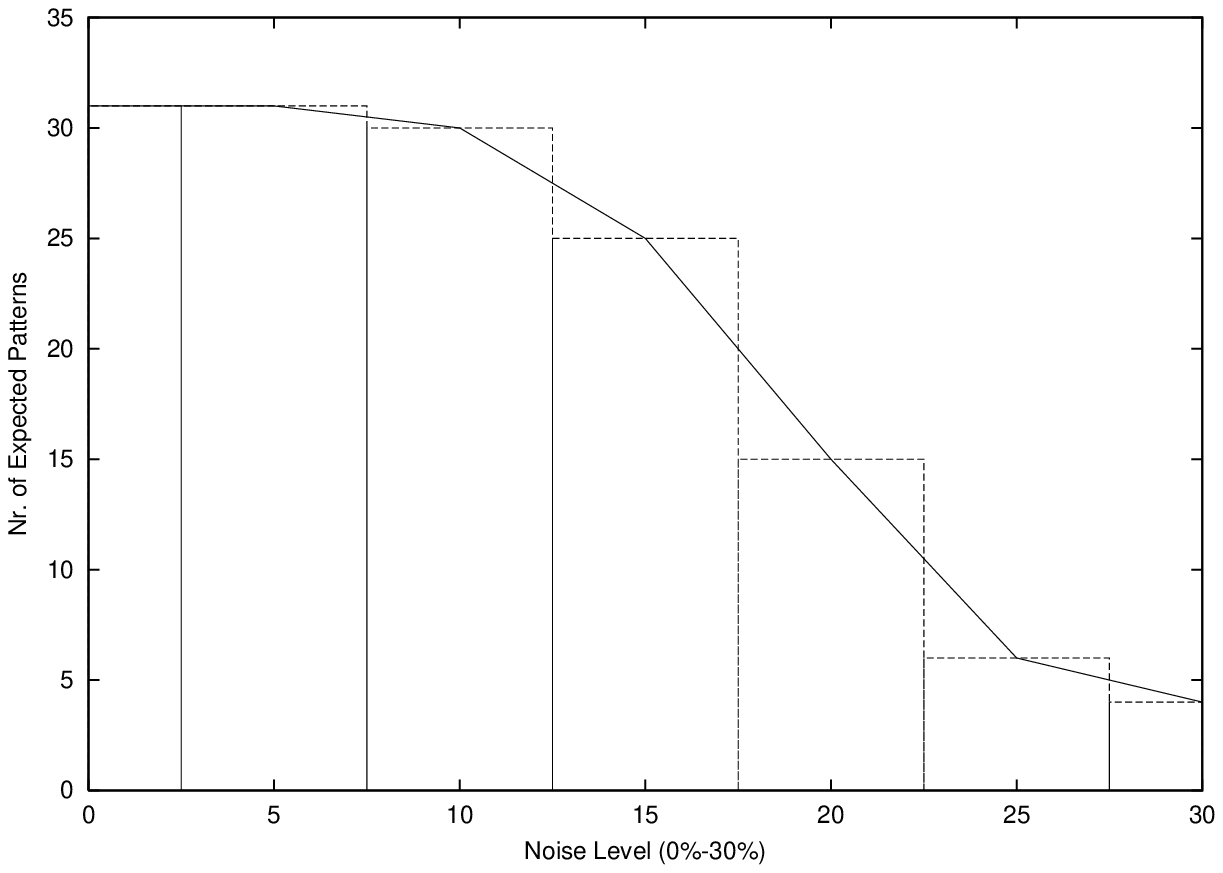}
\caption{The effect of noise on the algorithm.}
\label{fig:exp1}
\end{center}
\end{minipage}
\hfill
\begin{minipage}[t]{.49\textwidth}
\begin{center}
\includegraphics[width=6.0cm]{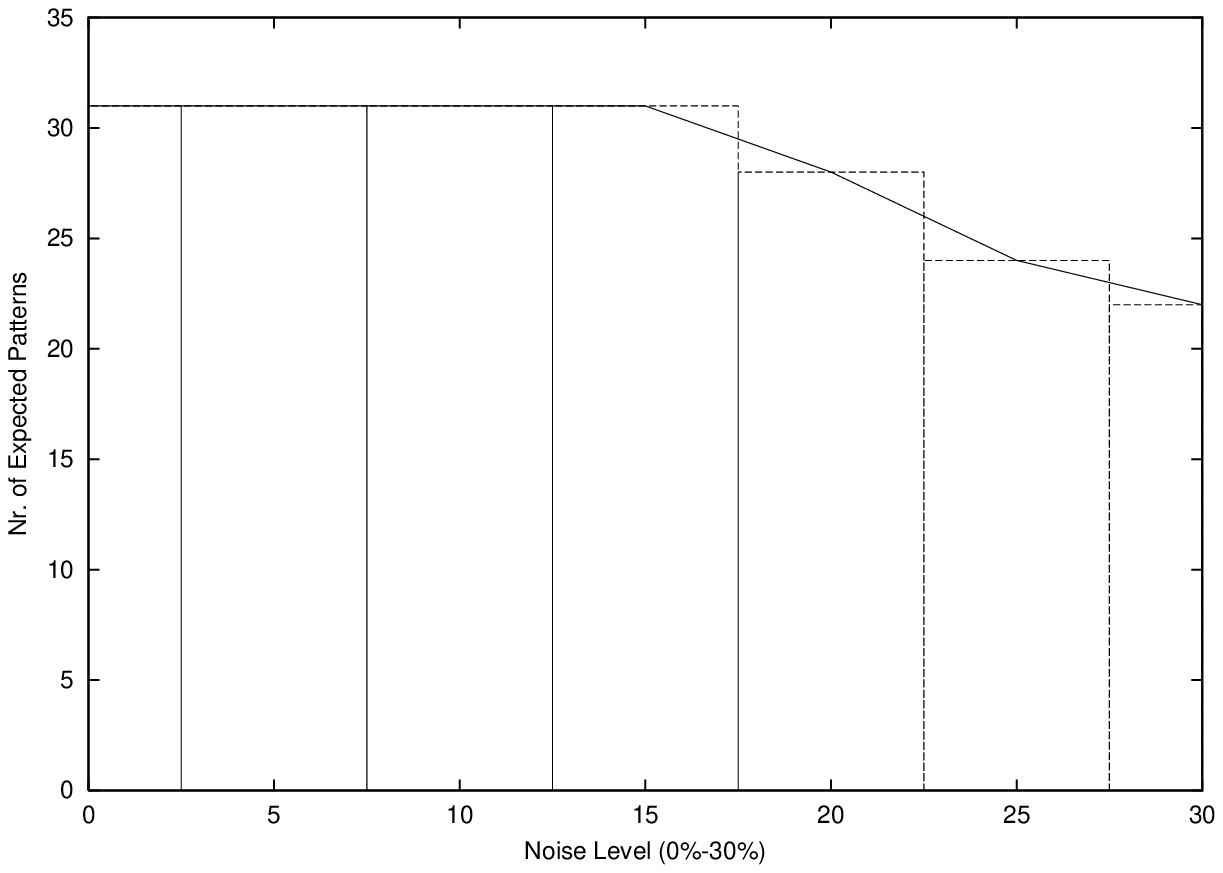}
\caption{The effect of noise on the algorithm, $\mathit{mindistfreq} = 50$.}
\label{fig:exp2}
\end{center}
\end{minipage}
\hfill
\end{figure}

The synthetic datasets used in our first experiment are called \emph{find-noise-x\%} 
where $x$ is a noise value ranging from 0 to 30. E.g., if the noise is $10\%$, 
this means there is a 10\% chance that one element of the balanced pattern does not occur
when it should. In each of these \emph{find-noise-x\%}  datasets one pattern of 5 of the 200 items occur
every 4 transactions (so distance = 3) and each dataset has 2,000 transactions. If 5 items always
occur balanced like this, we expect to find $\sum_{k=1}^{5}{5!}/{(5 - k)!k!} = 31$ patterns. First the 
\textsc{BalanceClat} algorithm is executed with $\mathit{maxstdev} = 2.5$, $\mathit{minavg} = 2.0$
and $\mathit{minnumber} = 150$. Figure \ref{fig:exp1} displays the number of expected patterns that were
found by the algorithm. We see that the algorithm detects most patterns up to a noise level of $15\%$. Due to 
the way we generate noise, long patterns become less likely as the noise level increases. With a high noise
level we only find the patterns of 1 item in length. This can be improved if we change our settings for 
$\mathit{maxstdev}$ and $\mathit{minavg}$, but we kept them fixed for comparison reasons.

We can use the $\mathit{mindistfreq}$ threshold to decrease the influence of small noisy periods on the
balanced occurrences. Figure \ref{fig:exp2} shows how the effect of noise becomes less if we set
a $\mathit{mindistfreq}$ of 50. Now one also finds more of the other patterns that happen to occur reasonably
balanced, however we can filter them by lowering $\mathit{maxstdev}$.

\begin{figure}[!ht]
\hfill
\begin{minipage}[t]{.49\textwidth}
\begin{center}
\includegraphics[width=6.0cm]{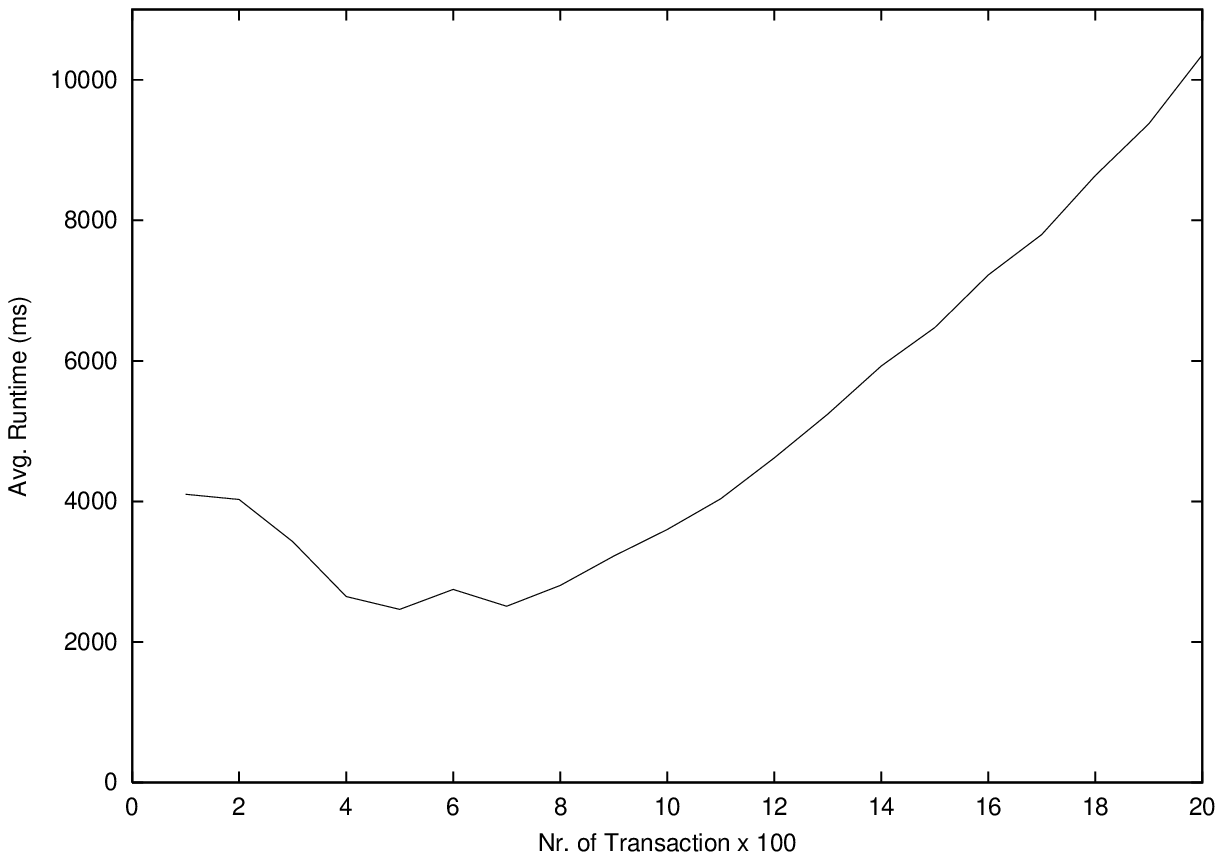}
\caption{Runtime in ms for different dataset sizes; $\mathit{minnumber}$ is 10\% of the dataset size ($\mathit{maxstdev} = 1.0$, $\mathit{minavg} = 2.0$, $\ell=10$).}
\label{fig:exp3}
\end{center}
\end{minipage}
\hfill
\begin{minipage}[t]{.49\textwidth}
\begin{center}
\includegraphics[width=6.0cm]{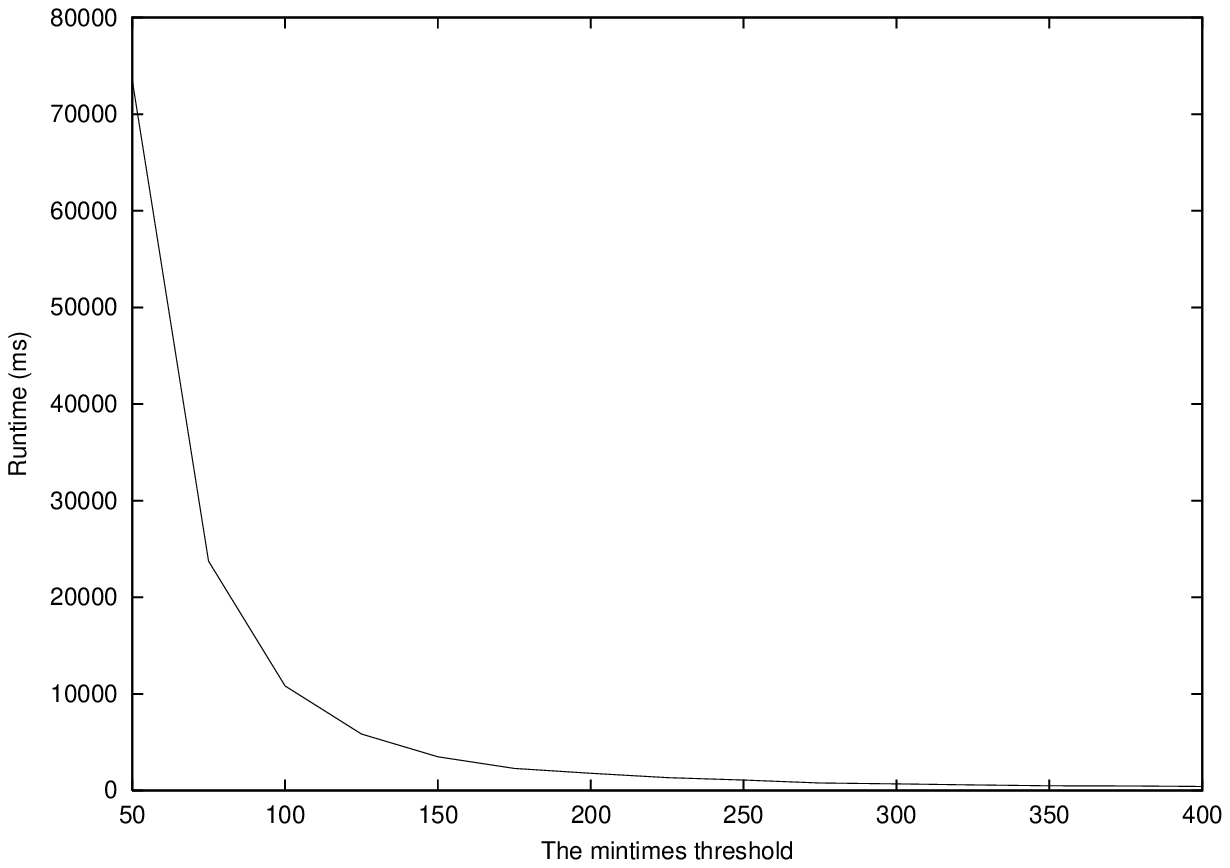}
\caption{Runtime in ms for different values of $\mathit{minnumber}$ ($\mathit{maxstdev} = 1.0$, $\mathit{minavg} = 2.0$, $\ell=10$).}
\label{fig:exp4}
\end{center}
\end{minipage}
\hfill
\end{figure}

With our next experiment we want to show the effect of dataset size on the algorithm, scalability. 
In Figure \ref{fig:exp3} first the runtime drops; this is because many patterns have distances occurring only a few times. E.g., when the
dataset size is 100 then $\mathit{minnumber} = 0.1 \cdot 100 = 10$. Many patterns have distances that occur at least
10 times. As this effect becomes less, runtime increases and eventually it becomes nearly linear. 

The \textsc{BalanceClat} algorithm was also tested on the website dataset. This dataset is based on an access log of the website of
the Computer Science department of Leiden University,
as said before. It contains all 1,991 items of the web-pages that were visited,
grouped in half-hour blocks, so each of the 1,488 transactions contains the pages visited during one half-hour. 
Figure \ref{fig:exp4} shows how the runtime for the website dataset drops fast as $\mathit{minnumber}$ increases.

Table \ref{table:occur} shows the count for distances between successive occurrences. It shows that
this particular pattern, consisting of the websites of two professors of the same group and the main page,
occurs often with a successive distance of 0, 1 or 2. This pattern probably is caused by students having
courses from both professors and some of these students access both pages nearly every half an hour.

\begin{table}
\begin{center}
\begin{tabular}{|c|c|}
\hline
In-between Transactions & Count \\
(Distance) & \\
\hline
0 & 385 \\
\hline
1 & 171 \\
\hline
2 & 78 \\
\hline
3 & 25 \\
\hline
4 & 23 \\
\hline
\end{tabular}
\end{center}
\caption{The distances (with count $\geq 20$) between successive occurrences and their counts 
for one pattern (two professors \& the main page) in the website dataset ($\mathit{maxstdev} = 2.0$, $\mathit{minavg} = 1.0$, $\ell = 10$).}\label{table:occur}
\vspace{-8mm}
\end{table}

Finally we also applied the \textsc{BalanceClat} algorithm to the Nakao dataset used in \cite{GraafGraaf}.
In this dataset each of the 2,124 transactions is a clone located on the human chromosomes. The items are the 
numbers of patients with a higher than normal value for this clone ($\geq 0.225$). The specifics of the dataset can be found
in \cite{NakaoK}. The parameter $\mathit{minavg}$ was set $0.0$, because the interesting patterns are expected
to occur very close to each other. Also $\mathit{mindistfreq} = 10$ because patterns where expected to have
small periods of transactions where they occurred unbalanced. 
Furthermore  $\mathit{maxstdev} = 0.2$, $\ell = 10$ and $\mathit{minnumber} = 100$. Results where similar to
results found with consecutive support as presented in \cite{GraafGraaf} where most consecutive patterns
occurred close together in chromosome 9. In the future we plan to investigate this futher.

\section{Conclusions and Future Work}

We have presented a new way of mining for patterns occurring with a regular interval.
In comparison with our previous method we now use a pruning threshold $\mathit{minnumber}$
that is more intuitive to users. With it the user only indicates the number of times at least
one intermediate distance should occur. Such a distance is the number of transactions between two occurrences
of the pattern (we consider only distances below a maximal distance).

In this work we call patterns with a regular interval balanced and we discuss an algorithm to
find them efficiently. Its runtime performance and scalability is evaluated through experimentation.

Finally in the future we plan to use balanced patterns in combination with new ways of filtering to
facilitate the discovery of new patterns further. Also research will be done on effectively
visualizing balanced patterns.

\bigskip 

\noindent \textbf{Acknowledgments}
This research is carried out within the Netherlands Organization for Scientific Research (NWO) MISTA Project (grant no. 612.066.304).

\end{document}